\documentclass[journal,twoside,web]{ieeecolor}
\usepackage{tmi}
\usepackage{cite}
\usepackage{amsmath,amssymb,amsfonts}
\usepackage{algorithmic}
\usepackage{graphicx}
\usepackage{textcomp}
\usepackage{stmaryrd}
\usepackage{booktabs}
\usepackage{multirow}
\usepackage{etoolbox}
\usepackage{hyperref}
\begin{document}
\title{PA-Seg: Learning from Point Annotations for 3D Medical Image Segmentation using Contextual Regularization and Cross Knowledge Distillation}
\author{Shuwei Zhai, Guotai Wang, Xiangde Luo, Qiang Yue, Kang Li and Shaoting Zhang
\thanks{This work was  supported by the National Natural Science Foundation of China (61901084, 62271115), National Key Research and Development Program of China (2020YFB1711500) and the 1·3·5 project for disciplines of excellence, West China Hospital, Sichuan University (ZYYC21004). (Corresponding author: Guotai Wang)}
\thanks{
S. Zhai, G. Wang, X. Luo and S. Zhang are with the School of Mechanical and Electrical Engineering, University of Electronic Science and Technology of China, Chengdu, 611731, China.  G. Wang, X. Luo and S. Zhang are also with Shanghai Artificial Intelligence Laboratory, Shanghai, 200030, China 
(e-mail: guotai.wang@uestc.edu.cn).}
\thanks{Q. Yue is with Department of Radiology, West China Hospital, Sichuan University, Chengdu, 610041, China}
\thanks{K. Li is with West China Biomedical Big Data Center, West China Hospital, Sichuan University, Chengdu 610041, China}
\thanks{This work has been submitted to IEEE TMI for possible publication. Copyright may be transferred without notice, after which this version may no longer be accessible.}}

\maketitle

\begin{abstract}
	The success of Convolutional Neural Networks (CNNs) in 3D medical image segmentation relies on massive fully annotated 3D volumes for training that are time-consuming and labor-intensive to acquire. In this paper, we propose to annotate a segmentation target with only seven points in 3D medical images, and design a two-stage weakly supervised learning framework PA-Seg. In the first stage, we employ geodesic distance transform to expand the seed points to provide more supervision signal. To further deal with unannotated image regions during training, we propose two contextual regularization strategies, i.e., multi-view Conditional Random Field (mCRF) loss and Variance Minimization (VM) loss, where the first one encourages pixels with similar features to have consistent labels, and the second one minimizes the intensity variance for the segmented foreground and background, respectively. In the second stage, we use predictions obtained by the model pre-trained in the first stage as pseudo labels. To overcome noises in the pseudo labels, we introduce a Self and Cross Monitoring (SCM) strategy, which combines self-training with Cross Knowledge Distillation (CKD) between a primary model and an auxiliary model that learn from soft labels generated by each other. Experiments on public datasets for Vestibular Schwannoma (VS) segmentation and Brain Tumor Segmentation (BraTS) demonstrated that our model trained in the first stage outperformed existing state-of-the-art weakly supervised approaches by a large margin, and after using SCM for additional training, the model's performance was close to its fully supervised counterpart on the BraTS dataset. 
\end{abstract}

\begin{IEEEkeywords}
Medical image segmentation, weakly supervised learning, noisy label, knowledge distillation.
\end{IEEEkeywords}

\begin{figure}[htb]
	\centering
	\includegraphics[width=0.49\textwidth]{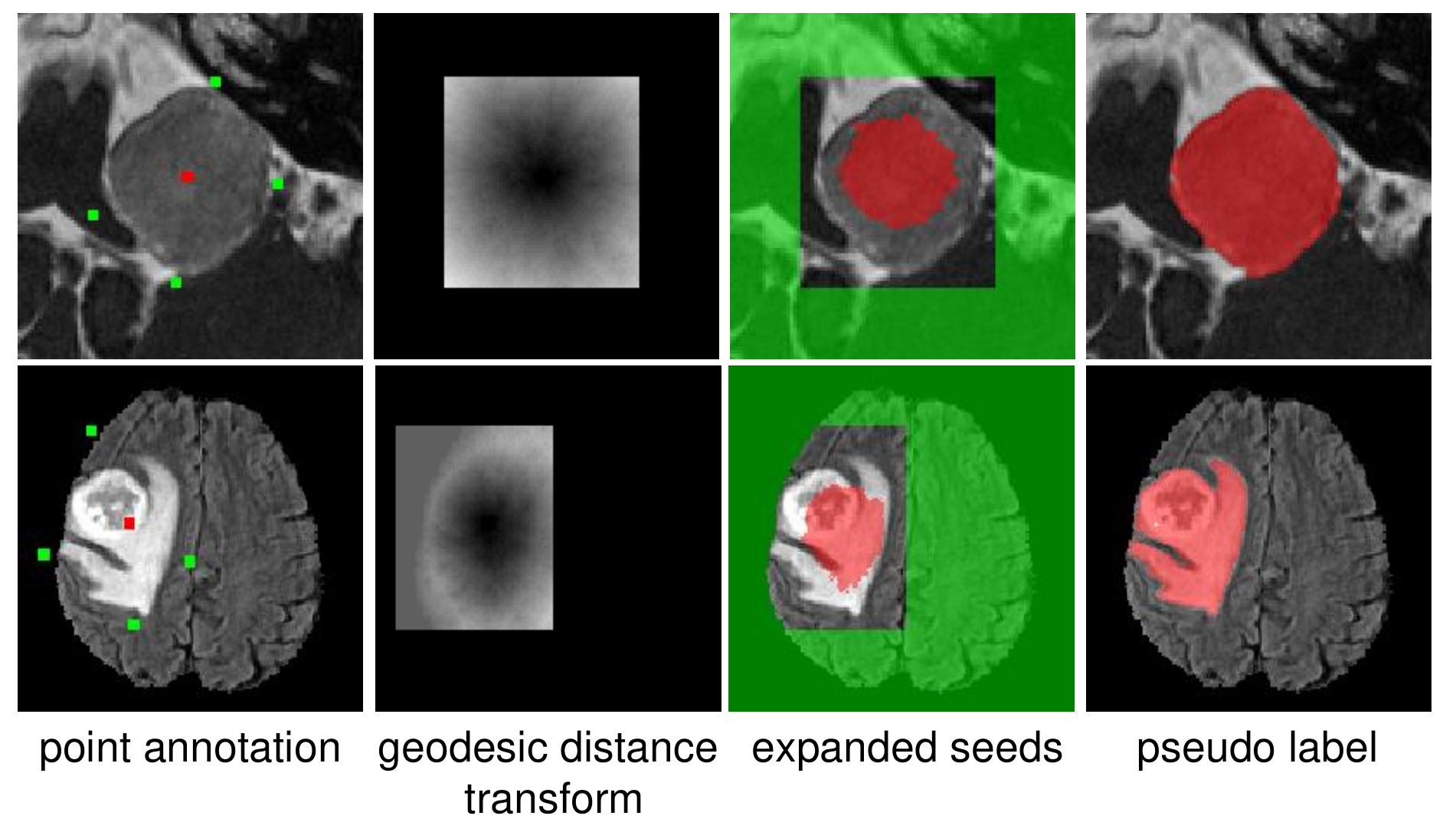}
	\caption{Illustration of our point annotation-based segmentation. Note that our method is implemented in 3D and the figure shows a 2D view. 
    The annotation for each object consists of six points in different axis directions in the background (four of them are shown in the 2D slice) and one in the foreground. The sparse annotations are expanded based on geodesic distance transform of the foreground seed and bounding box derived from the background seeds. Pseudo labels are generated by a model trained with the expanded annotations. } \label{fig:7points}
\end{figure}

\section{Introduction}
\label{sec:introduction}
\IEEEPARstart{M}{edical} image segmentation plays a vital role in computer-assisted  diagnosis and treatment planning of many diseases. In recent years, deep Convolutional Neural Networks (CNNs) have  achieved breakthrough performance for segmentation when fully annotated datasets are available \cite{litjens2017survey,hesamian2019deep}. 
However, their success relies on large-scale training images with dense annotations that are time-consuming to obtain and rely on experts with domain knowledge~\cite{weese2016four}. 

To reduce the annotation cost for medical image segmentation, increasing efforts have been made to exploit weak or sparse annotations that are much more efficient and easier to obtain than dense pixel-level annotations, e.g., image-level annotations \cite{wang2020self,wu2019weakly}, scribbles \cite{lin2016scribblesup,lee2020scribble2label}, bounding boxes \cite{rajchl2016deepcut,kervadec2020bounding} and point annotations \cite{roth2021going,dorent2021inter}. 
Despite the efficiency of image-level annotations, their resulted segmentation performance is usually very limited due to the extremely weak supervision.
Scribbles are widely used for efficient annotation of 2D images~\cite{lin2016scribblesup,lee2020scribble2label}, and they provide labels for a sparse set of pixels of each class for training. For 3D volumes, scribble annotations need to be provided for many slices, which is still time-consuming.
In contrast, bounding box annotations~\cite{rajchl2016deepcut,kervadec2020bounding} are cheaper than scribbles and can provide coarse location of the target for training a better segmentation model than image-level annotations. However, such annotations are faced with insufficient labeled foreground pixels for training, 
and cannot accurately approximate the target boundary for complex shapes, 
which will misguide the segmentation model. Annotation with few points has a much lower cost than scribbles, and can better deal with complex shapes compared with bounding box annotations, thus has a potential for reducing the annotation cost while achieving good performance. For example, some researchers \cite{roth2021going,dorent2021inter}  used  extreme points for annotating 3D objects, where some paths between extreme points are used as foreground annotations for training a segmentation model. However, the exact extreme points in 3D space are hard to find, especially for 3D medical images with a low contrast and tumors with irregular shapes, which increases the time cost of annotators and may result in inaccurate extreme points that limit the training process.

Despite its efficiency for annotation, there are two main challenges for learning from weak annotations for segmentation of 3D objects. First, due to the large number of unannotated pixels, the segmentation model tends to over-fit the small number of annotated pixels if any. To tackle this problem, some works utilize prior information for regularization, such as size constraint~\cite{kervadec2019constrained} 
and tightness prior~\cite{kervadec2020bounding}. Such regularization strategies help to improve the model's performance, but their applicability is still subject to certain assumptions, e.g., the segmentation target has a regular shape. Second, pseudo dense annotations may be generated from the weak annotations, but they usually contain errors (noises) that will cause severe performance degradation~\cite{tang2018normalized,tang2018regularized}. Several approaches have been proposed to deal with this problem\cite{zhang2018generalized, wang2020noise}, 
and most of them assume samples with high loss values are likely to be noisy and select those with low loss values for robust learning. However, it is difficult to select a reliable loss threshold, and these methods can hardly distinguish noisy samples from  hard samples, which limits the model's performance.

In this work, we propose a novel point annotation-based weakly supervised method named as PA-Seg for 3D medical image segmentation. It only requires seven points (seeds) to annotate a 3D segmentation target, i.e., six points on the background in different axis directions (left, right, anterior, posterior, inferior and superior) and one point on the foreground. An example of the axial view of our annotation is shown in the first column of Fig.~\ref{fig:7points}. Compared with bounding box annotation, our annotation provides extra foreground information at the cost of only one additional mouse click. In addition, providing the background points is more friendly and easier  than finding the extreme points for annotators.  

The training process of PA-Seg consists of two stages: the first stage uses the sparse point annotations to train an initial segmentation model that generates dense pseudo labels for training images, and the second stage trains an enhanced model based on the pseudo labels and noise-robust learning. 
A challenge in the first stage is that the seven points cannot be directly used for training due to the insufficient supervision signal. To deal with this problem, we utilize geodesic distance transform \cite{wang2018deepigeos,luo2021mideepseg} to expand the raw annotation seeds to derive the annotations of more pixels, as shown in the third column of Fig.~\ref{fig:7points}. The expanded annotations with fewer unannotated pixels are used to train the initial segmentation model, where we introduce contextual regularization on the pixels with unknown labels based on a multi-view Conditional Random Field (mCRF) loss and a Variance Minimization (VM) loss. 
Training with our seed expansion strategy and regularization methods can obtain a high-performance initial segmentation model that is superior to several state-of-the-art weakly supervised methods. The initial model generates high-quality pseudo labels for training images, as shown in the last column of Fig.~\ref{fig:7points}.

In the second stage, to further learn from the noisy pseudo labels, we introduce an auxiliary network to assist training of the primary network, and both of them are pre-trained by the method in the first stage. They are then trained by our proposed Self and Cross Monitoring (SCM) that combines self-training and Cross Knowledge Distillation (CKD). The CKD is inspired by Knowledge Distillation (KD)~\cite{hinton2015distilling}, where soft labels predicted by a teacher model contain more information than the hard labels that may contain noises. Differently from  typical KD methods \cite{hinton2015distilling,sarfraz2021knowledge} that focus on image classification problems and only distill the knowledge from the teacher to the student, our CKD lets two networks distill knowledge from each other for image segmentation, where the different decision boundaries of two models can regularize each other and reduce the risk of over-fitting in a single model. 

\subsection{Contributions}
The contributions of this paper are three-fold. Firstly, we propose a novel framework PA-Seg for weakly supervised 3D medical image segmentation, where only seven points are needed to annotate a 3D object, and a two-stage method based on pseudo label generation and learning from noisy pseudo labels is proposed for training a segmentation model with high performance and very low annotation cost.
Secondly, to deal with the extremely sparse annotation seeds, we propose a geodesic distance-based  annotation expansion strategy, and introduce contextual regularization based on multi-view conditional random field and variance minimization to train a good initial segmentation model. Thirdly, we propose self and cross monitoring that combines self-training and cross knowledge distillation for learning from noisy pseudo labels in the scenario of weakly supervised segmentation. Extensive experiments on two public datasets for Vestibular Schwannoma (VS) segmentation and Brain Tumor Segmentation (BraTS) demonstrated that our method outperformed several state-of-the-art weakly supervised techniques, and it was comparable to fully supervised setting on the BraTS dataset. 

\section{Related Works}
\subsection{Weakly Supervised Segmentation Using CNNs}
Existing weakly supervised CNNs for segmentation can be roughly divided into two categories. The first category of methods use traditional interactive segmentation techniques (e.g, Grabcut~\cite{rother2004grabcut} and random walker~\cite{grady2006random}) to produce ``fake" dense masks (i.e., pseudo labels) from the given weak annotations like bounding boxes and scribbles. These pseudo labels are then used to train CNNs in a fully-supervised manner~\cite{rajchl2016deepcut,lin2016scribblesup}. For example, ScribbleSup~\cite{lin2016scribblesup} alternatively propagates scribble information to unlabeled pixels and optimizes the network parameters. 
However, the pseudo labels are usually not correct and contain a lot of noises, and standard segmentation loss functions such as cross-entropy and Dice losses may over-fit the noises, leading to limited performance \cite{wang2020noise,zhang2018generalized}. 

The second category of methods combine partial cross entropy loss   for  annotated pixels 
with regularization for unannotated pixels.  Kervadec~\cite{kervadec2019constrained} used size loss  to constrain volume of the predicted mask, and employed deep tightness prior \cite{kervadec2020bounding} for learning 
from bounding box annotations. 
Tang et al. \cite{tang2018normalized,tang2018regularized} proposed normalized cut loss and Conditional Random Field (CRF) loss to regularize unannotated pixels. 
However, CRF loss involves pairwise energy for any pair of pixels in the image, which is not suitable for 3D images due to the huge memory and computational cost. 

\subsection{Co-learning Framework}
Co-learning is a kind of framework where two networks or two branches of a single network are used for training with imperfect annotations, and it has been recently used for semi-supervised~\cite{yu2019uncertainty,chen2021semi}, weakly-supervised~\cite{luo2022scribble}, and noise-robust learning~\cite{han2018co}.
For semi-supervised learning, the Mean Teacher (MT) framework~\cite{tarvainen2017mean,yu2019uncertainty} utilized
an exponential moving average of a student as the teacher to provide supervision signal on unannotated images.  
Wang et al.~\cite{wang2021self} let two or more models focus on easier-to-segment regions first, and then gradually consider harder ones, while enforcing predictions from different models to be consistent. Zhao et al.~\cite{zhao2021mt} applied MT to unsupervised domain adaptation, where the student learns from two teacher models each from a different domain in a semi-supervised manner. The Cross Pseudo Supervision (CPS)~\cite{chen2021semi} lets two networks generate pseudo labels for each other in semi-supervised learning.

For weakly-supervised learning,  Liu et al.~\cite{liu2021mixed} used a strongly supervised branch to teach a less supervised branch. Luo et al.~\cite{luo2022scribble} employed a dynamic mixture of predictions from two branches as the supervision.  Patel et al.~\cite{patel2022weakly} proposed cross-modality equivariant constraints for weakly supervised segmentation. 
For noise-robust learning, Co-teaching~\cite{han2018co} uses two networks  each selects samples with small loss values to supervise the other. Xue et al.~\cite{xue2020cascaded} extended co-teaching~\cite{han2018co} by using three networks to select and correct noisy samples for robust learning. 
Wei et al.~\cite{wei2020combating} proposed to use symmetric Kullback-Leibler divergence in addition to cross-entropy loss to select small-loss instances. However, most existing works use two or more networks with the same architecture for co-learning. 
The asymmetry design is important for improving the performance of co-learning. Existing works have designed  asymmetry based on different modalities~\cite{patel2022weakly}, processing inputs with different noises or augmentation~\cite{ tarvainen2017mean,yu2019uncertainty}, using different objectives~\cite{Shi2022tmi} or prediction representations~\cite{luo2021semi}. In this work, we introduce architecture-level asymmetry, i.e,  use two networks with different architectures to avoid feature bias of a single network, and combine them with cross knowledge distillation to better deal with noisy pseudo labels. 

\subsection{Learning from Noisy Labels and Knowledge Distillation}
Noisy labels exist in either inaccurate manual annotations or pseudo masks generated by a trained model. Due to the strong fitting ability of CNNs, training with such labels using standard fully supervised learning will over-fit the noises and  limit the performance of the segmentation model~\cite{karimi2020deep}. 
Existing noise-robust learning methods can be roughly grouped into two categories. The first is label filtering and sample re-weighting by using two or multiple branches as mentioned above~\cite{han2018co,xue2020cascaded,wei2020combating}. 
The second is to design noise-robust loss functions, such as Mean Absolute Error (MAE) loss~\cite{ghosh2017robust}, Generalized Cross-Entropy (GCE) loss~\cite{zhang2018generalized} and noise-robust Dice loss~\cite{wang2020noise}. 

Knowledge Distillation (KD)~\cite{hinton2015distilling} was originally proposed for transferring  knowledge from a complex teacher model to a lightweight student.  In medical imaging, KD has been applied to different tasks where the knowledge is distilled from one model/branch to improve the performance of another.  Wang et al.~\cite{wang2019segmenting} used KD to boost a lightweight student network for computationally efficient segmentation of 3D volumes. Patel et al. ~\cite{patel2022weakly} used cross-modality KD to improve the accuracy of Class Activation Maps (CAMs) for learning from image-level labels.  Dou et al.~\cite{dou2020unpaired} employed KD to  explicitly  constrain the KL-divergence of prediction distributions between two different modalities.

KD~\cite{hinton2015distilling}  typically uses soft labels 
 based on temperature scaling  instead of hard one-hot ones from the teacher to supervise the student. As soft labels contain more information about the relationship between different classes, they can lead to improved performance of the student compared with only learning from standard one-hot hard labels that may contain noises.
 A similar conclusion was found in~\cite{lukasik2020does} where label smoothing helps to improve the model's performance when the labels are inaccurate. 
 Kats et al.~\cite{kats2019soft} found that soft labels constructed by morphological dilation of binary annotation masks improved the performance of brain lesion segmentation.
Sarfraz et al.~\cite{sarfraz2021knowledge} showed the effectiveness of KD in learning efficiently under various degrees of label noise and class imbalance. Li et al.~\cite{li2017learning} leveraged the knowledge learned from a small clean dataset to guide the learning from a noisy dataset. However, such a clean dataset is not available in weakly supervised setting. In addition, most existing methods using KD for noise-robust learning are proposed for image classification tasks, and their applicability to 3D medical image segmentation has not been well investigated.  

\begin{figure*}[htb]
	\centering
	\includegraphics[width=\textwidth]{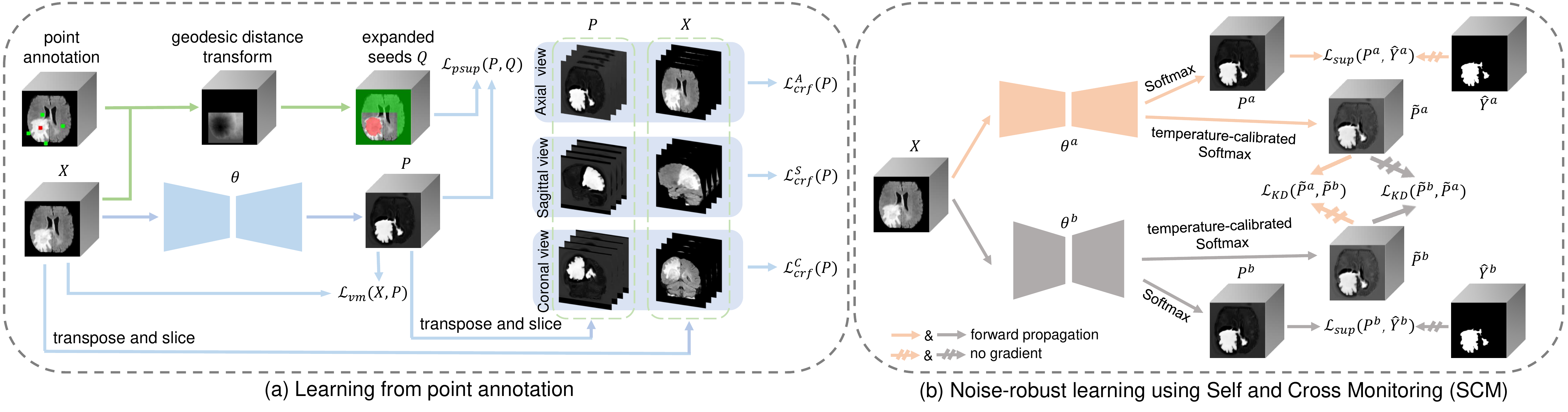}
	\caption{An overview of the proposed PA-Seg for weakly supervised 3D segmentation based on point annotations. In the first stage (a), we expand the annotation seeds based on geodesic distance transform, and train an initial model using the expanded seeds, with the unlabeled pixels regularized by multi-view CRF loss and Variance Minimization (VM) loss. Pseudo labels are then obtained by using the initial model for inference.  In the second stage (b), to deal with noises in the pseudo labels, we propose Self and Cross Monitoring (SCM), where a primary model and an auxiliary model supervise each other via Cross Knowledge Distillation (CKD) based on soft labels, in addition to self-training of each model.}
	\label{fig:overview}
\end{figure*}
\section{Method}
Fig.~\ref{fig:overview} shows an overview of our proposed PA-Seg for weakly supervised 3D segmentation  using seven points for annotation (six on the background and one on the foreground). In the first stage, we utilize geodesic distance transform to expand the extremely sparse seed points, and combine multi-view Conditional Random Field (mCRF) loss with Variance Minimization (VM) loss to train an initial segmentation model that can produce high-quality pseudo labels for training images. In the second stage, the segmentation model is further boosted by learning from the pseudo labels. We design a novel procedure named Self and Cross Monitoring (SCM) that combines self-training and Cross Knowledge Distillation (CKD) to alleviate the effect of noises in the pseudo labels and obtain better segmentation results. 
\subsection{Learning from Point Annotation}\label{sec:method:segmentation}
\subsubsection{Annotation Strategy}
The first column of Fig.~\ref{fig:7points} shows our proposed annotation strategy based on points. Note that the figure shows a 2D view and our method works in 3D. For a 3D object to annotate, the annotator just needs to provide seven points, i.e., one inside the target object as foreground seed and six outside the target as background seeds. The foreground seed is provided near the center of the object, but the annotator does not need to find the exact centroid for convenience. The six background seeds are provided in different axis directions, i.e., the left, right, anterior, posterior, inferior and superior side out of the object. Note that each background seed should be away from the corresponding extreme point, e.g., the x-axis coordinate of the left/right seed should be smaller/larger than that of the most left/right point in the object. The background seeds are so provided that a bounding box determined by them will cover the entire target object. This requirement can be easily satisfied when the annotator estimates the extreme points roughly and expands them to some extent for clicking.

\begin{figure}
	\centering
	\includegraphics[width=0.49\textwidth]{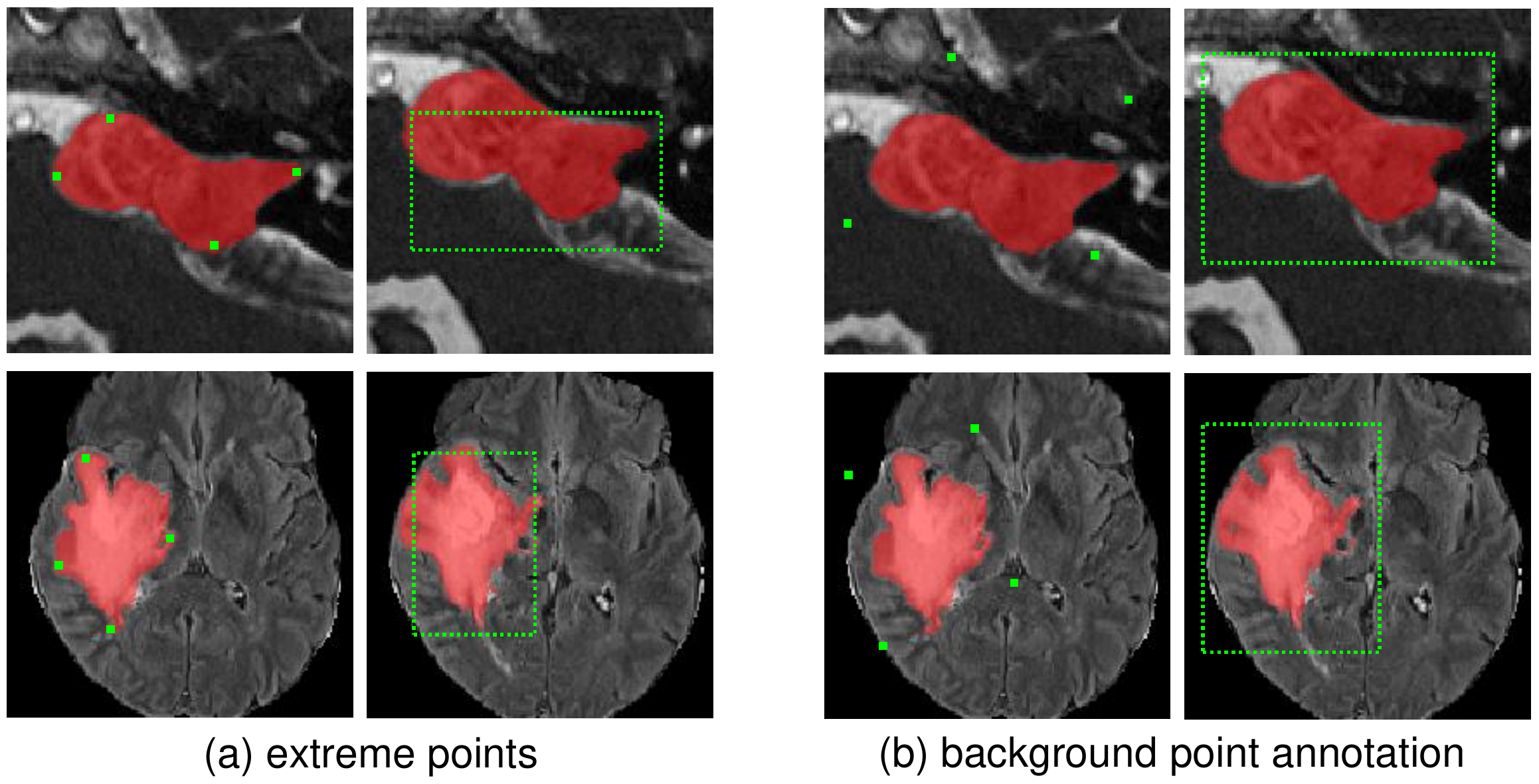}
	\caption{Comparison between extreme points and background points for annotation. The two rows are for VS and glioma, respectively. The green points are annotations, and the dotted lines are the corresponding bounding boxes overlaid on nearby slices.}
	\label{fig:anno_strategy}
\end{figure}

Compared with providing extreme points~\cite{roth2021going,dorent2021inter}, our annotation strategy allows more freedom for the annotator who does not need to find the exact 3D extreme points, which leads to higher efficiency and convenience for annotating. 
In  3D medical images, it is very difficult to find the accurate positions of extreme points in 3D space, especially for tumors with irregular shapes. As shown in  Fig.~\ref{fig:anno_strategy} (a), the extreme points identified in one slice by the annotator may not be those for the 3D object, which leads to underestimated bounding boxes that only cover a sub-region of the 3D object. In contrast, as shown in  Fig.~\ref{fig:anno_strategy} (b), our method does not require the user to provide exact extreme points, but just some relaxed points outside the visually estimated bounding box. A bounding boxed derived from these points will be a superset of the real bounding box, which can tolerate inaccurate points and ease the annotation. 

\subsubsection{Seed Expansion}\label{sec:method:expansion}
As only seven pixels are labeled in a 3D image, they cannot be directly used to train a segmentation model due to the insufficient supervision signal compared with millions of parameters in the CNN. Therefore, we expand the seeds to reduce the number of unlabeled pixels automatically before training. 

Firstly, based on our annotation strategy, pixels outside the bounding box determined by the background seeds can be safely considered as the background region. Let $\Omega$ denote the entire image space, $\Omega_{in}$ and $\Omega_{out}$ denote the regions inside and the outside the bounding box, respectively. Note that in some segmentation tasks, the input image may contain some air background or be prepossessed by removing some undesired regions like the skull in the BraTS dataset, leading to zero values around the valid image region (e.g., the brain), as shown in the second row of Fig.~\ref{fig:7points}. Let $\Omega_{M}$ denote the masked valid region that excludes the undesired background. We therefore take a union of $\Omega_{out}$ and $\Omega - \Omega_{M}$ as the set of background pixels: $\Omega_B = \Omega_{out}\cup (\Omega - \Omega_{M})$.

Secondly, the foreground seed is expanded to some degree to obtain more foreground pixels with the assumption that pixels close to the foreground seed with a similar intensity are likely to be foreground as well. As Euclidean distance does not contain contextual information and cannot deal with objects with non-ellipsoid  shapes, geodesic distance~\cite{wang2018deepigeos,luo2021mideepseg} is used to deal with this problem since it can distinguish adjacent pixels with different appearances and improves label consistency in homogeneous regions. Let $X$ and $i_f$ denote a training image and the foreground seed, respectively. For a pixel in $i \in \Omega - \Omega_B$, the geodesic distance between $i$ and $i_f$ in $X$ is: 
\begin{equation}\label{eq:geodesic}
	\mathcal{D}(i,i_f) = \min_{p\in\mathcal{P}_{i,i_f}}{\int_{0}^{1}|| \nabla{X}(p(n))\cdot\mathbf{v}(n)|| dn}
\end{equation}
where $\mathcal{P}_{i,i_f}$ is the set of all possible paths between $i$ and $i_f$. $p$ is one feasible path and parameterized by $n\in [0, 1]$. $\mathbf{v}(n)$ is a unit vector tangent to the direction of the path at $p(n)$. The maximal geodesic distance value in $\Omega - \Omega_B$ is denoted as $\mathcal{D}_{max}$. We use $\Omega_F$ to denote the set of expanded foreground seeds, and it is obtained by:
\begin{equation}\label{eq:foreground_seed}
	\Omega_F = \{i|i\in \Omega -\Omega_B ~\text{and}~ \mathcal{D}(i,i_f) < \epsilon \mathcal{D}_{max} \}
\end{equation}
where $\epsilon \in [0, 1]$ is a hyper-parameter to control the degree of seed expansion. $\epsilon=0$ means no expansion, and $\epsilon=1$ means expansion to the entire region of $\Omega - \Omega_{B}$. For simplicity, we denote the set of labeled pixels as $\Omega_L = \Omega_B \cup \Omega_F$, and denote the set of pixels with unknown labels as $\Omega_{U} = \Omega - \Omega_L$. An example of the seed expansion result is shown in the third column of Fig.~\ref{fig:7points}, where the green and red colors show $\Omega_B$ and $\Omega_F$, respectively, and the remaining pixels belong to $\Omega_U$.

\subsubsection{Partially Supervised Loss}\label{sec:method:partially}
After seed expansion, we obtain a partial label map $Q$ where values for pixels in $\Omega_{F}$ and $\Omega_{B}$ are 1 and 0, respectively. Then $Q$ can be used to train a segmentation network via a partially supervised loss:
\begin{equation}\label{eq:pSup}
	\mathcal{L}_{psup}(P,Q) = \big( \mathcal{L}_{pce}(P, Q) + \mathcal{L}_{pdc}(P, Q) \big)/2
\end{equation}
Where $P$ is the foreground probability map predicted by the segmentation model. $\mathcal{L}_{pce}$ and $\mathcal{L}_{pdc}$ are partial cross-entropy and partial Dice losses, respectively:
\begin{equation}\label{eq:pCE}
	\mathcal{L}_{pce}(P, Q) = \frac{1}{|\Omega_L|}\sum_{i \in \Omega_L} -\big( q_i \log(p_i) + (1 - q_i) \log(1 - p_i) \big)
\end{equation}
\begin{equation}\label{eq:pDC}
	\mathcal{L}_{pdc}(P,Q) = 1 - \frac{2 \sum_{i \in \Omega_L}{p_iq_i}}{\sum_{i \in \Omega_L}({p_i}^2 + {q_i}^2)}
\end{equation}
where $p_i$ is the foreground probability of pixel $i$ in $P$, and $q_i$ is the corresponding value in $Q$. Here we follow the common practice of combing the cross entropy loss with Dice loss for labeled pixels~\cite{zhao2021mt}, where the former has a large gradient for each pixel, and the latter can better deal with the class imbalance problem for segmentation tasks. Note that in Eq.~\ref{eq:pCE} and Eq.~\ref{eq:pDC}, the loss values are only calculated in $\Omega_L$, and pixels with unknown labels are ignored.

\subsubsection{Multi-view CRF Loss}\label{sec:method:mCRF}
Dense CRF has been widely used for post-processing in segmentation due to its ability to capture long-range dependency. In dense CRF, a pairwise potential is defined between two different pixels in a 2D image to encourage consistent labels between pixels with similar features:

\begin{equation}
	\psi(y_i,y_j) = \mu(y_i,y_j)K_{ij},~K_{ij} = \sum_{n=1}^N w_n k_n(\mathbf{f}^{(n)}_i, \mathbf{f}^{(n)}_j) 
\end{equation}
where $y_i$ is the label of pixel $i$. $\mu$ is a class compatibility matrix based on the Potts model, i.e., $\mu(y_i,y_j)=1$ if $y_i\neq y_j$ and 0 otherwise. $K_{ij}$ is the discontinuity cost based on a mixture of $N$ kernels. $w_n$ is the weight of the $n$-th kernel $k_n$ that is commonly implemented by a Gaussian kernel~\cite{Krah2011}, and the feature vector $\mathbf{f}^{(n)}_i$ is specific to $k_n$, such as pixel intensity and spatial coordinate used in the bilateral filter. 
Following~\cite{tang2018regularized}, the Potts model can be relaxed from hard $y_i$ to soft $p_i$. A quadratic relaxation leads to the 2D dense CRF loss:
\begin{equation}\label{eq:CRFloss}
	\mathcal{L}_{crf}(P^{(s)}) = \sum_{i, j\in \Omega^{(s)}} p_i (1-p_j)K_{ij}
\end{equation}
where $P^{(s)}$ is the $s$-th slice of a 3D prediction map $P$, and $\Omega^{(s)}$ denotes the set of pixels in that slice.

The CRF loss can be directly used to regularize the network's prediction during training, which is more efficient than using dense CRF for post-processing after each round of training~\cite{rajchl2016deepcut}.

However, as the CRF loss involves $K_{ij}$ for any pair of pixels in the image, it leads to huge computational and memory cost, which is not suitable for 3D images. Assume the 3D image has a shape of $D\times H \times W$, then the shape of the affinity matrix $K=[K_{ij}]$ will be $(DHW)\times (DHW)$, which is inefficient and even impossible to compute for large 3D volumes.  To enable 3D contextual regularization and avoid the computational problem, we introduce multi-view CRF loss that applies the CRF loss in axial, sagittal and coronal views, respectively. The axial CRF loss for a 3D volume is defined as the average CRF loss for the $D$ slices:  $\mathcal{L}^A_{crf}=\big( \sum_{d}\mathcal{L}_{crf}(P^{(d)}) \big) / D$. Similarly, the sagittal and coronal CRF losses are denoted as $\mathcal{L}^S_{crf}$ and  $\mathcal{L}^C_{crf}$, respectively, and they are the average 2D CRF loss across the $W$ sagittal and $H$ coronal slices, respectively. The proposed multi-view CRF loss is:
\begin{equation}\label{eq:mCRF}
		\mathcal{L}_{mcrf}(P) = \frac{1}{3}\big(	\mathcal{L}^A_{crf}(P) + \mathcal{L}^S_{crf}(P)
		+ \mathcal{L}^C_{crf}(P)\big)
\end{equation}

\subsubsection{Variance Minimization}
Motivated by the Chan-Vese level set model~\cite{chan2001active} for unsupervised segmentation, we use a similar formulation 
to regularize the segmentation by minimizing the intensity variance in the segmented foreground and background regions, respectively. Here we denote the average intensities for the segmented foreground and background as $u_f$ and $u_b$, respectively. They are obtained by:
\begin{equation}\label{eq:mean}
	u_f = \frac{\sum_{i \in \Omega_M} {x_ip_i}}{\sum_{i \in \Omega_M} {p_i}},~~u_b = \frac{\sum_{i \in \Omega_M} {x_i(1 - p_i)}}{\sum_{i \in \Omega_M} {(1- p_i)}}
\end{equation}
where $x_i$ is the intensity  of pixel $i$ in the input image. Note that we only calculate the average across the valid image region $\Omega_M$ to exclude influence of  the undesired image region (e.g., air background). The Variance Minimization (VM) loss is:

\begin{equation}\label{eq:VM}
	\begin{aligned}
		\mathcal{L}_{vm}(X, P) &= \frac{\sum_{i \in \Omega_M} {(x_i-u_f)^2 p_i}}{\sum_{i \in \Omega_M} {p_i}} \\
		&+ \frac{\sum_{i \in \Omega_M} {(x_i-u_b)^2(1 - p_i)}}{\sum_{i \in \Omega_M} {(1 - p_i)}}
	\end{aligned}
\end{equation}

\subsubsection{Overall Loss}
The overall loss for training the intial segmentation model in the first stage  is summarized as:
\begin{equation}\label{eq:overall-1}
	\begin{aligned}
		\mathcal{L}_{seg} &= 
		\mathcal{L}_{psup}(P,Q)
		+ \alpha\mathcal{L}_{mcrf}(P) + \beta \mathcal{L}_{vm}(X,P)
	\end{aligned}
\end{equation}
where $\alpha$ and $\beta$ are hyper-parameters that control the weights of the corresponding losses.
\subsection{Noise-robust Learning Using Self and Cross Monitoring}\label{sec:method:SCM}
The initial model trained in the first stage can generate pseudo labels for training images to train an enhanced model. To deal with noises in the pseudo labels, we design a novel Self and Cross Monitoring (SCM) method to improve the model's performance, as illustrated in Fig.~\ref{fig:overview}(b). The idea is inspired by Knowledge Distillation~\cite{hinton2015distilling}, where soft labels obtained by a teacher can enhance the student's performance. Differently from typical KD methods that fix the teacher and only update the student, we introduce Cross Knowledge Distillation (CKD) so that two models can teach each other with soft labels. The motivation is that two different networks have different decision boundaries, and an interaction between them can regularize each other and reduce the risk of over-fitting caused by the bias of a single network. 

Let $\theta^a$ denote a primary network and $\theta^b$ denote an auxiliary network, and they have been pre-trained independently in the first stage. Let $Z^a$ and $Z^b$ denote the logit map of the penultimate layer of the two networks, respectively. Passing $Z^a$ or $Z^b$ to a standard Softmax layer leads to a probability prediction, $P^a$ or $P^b$. For simplicity, we use $P^a$ and  $P^b$ to denote the foreground probability maps from the two networks, respectively. The corresponding hard labels based on $argmax(Z)$ for the two pre-trained networks are denoted as $\hat{Y}^a$ and $\hat{Y}^b$, respectively, i.e, pseudo labels. At the same time, a soft prediction is obtained by sending the logit map to a temperature-calibrated Softmax layer for KD~\cite{hinton2015distilling}:
\begin{equation}\label{eq:softmax}
	\tilde{p}_i = \frac{e^{z_{i,f}/T}}{e^{z_{i,f}/T} + e^{z_{i,b}/T}}
\end{equation}
where $z_{i,f}$ and $z_{i,b}$ are the logit scores for the foreground and background at pixel $i$, respectively. $\tilde{p}_i$ is the soft foreground probability at pixel $i$, and $T$ is the temperature parameter. Note that a standard Softmax corresponds to $T=1$, and a higher $T$ produces a softer probability output. As shown in previous works~\cite{guo2017calibration,muller2019does,Xu2020fnkd}, temperature scaling with $T>1$  plays an important role in reducing the effect of noise, as it avoids over-confidence of CNNs. In our task, the pseudo labels from the teacher may contain inaccurate parts that will misguide the student, softening the pseudo labels can avoid over-confidence of the teacher for noise-robust learning. 

Let $\tilde{P}^a$ and $\tilde{P}^b$ denote the ``soft" foreground probability maps of the two networks, respectively. When taking $\theta^a$  and $\theta^b$ as the student and the teacher respectively, the KD loss based on Kullback Leibler (KL) divergence is:
\begin{equation}\label{eq:kd}
	\mathcal{L}_{KD}(\tilde{P}^a,\tilde{P}^b) = \frac{1}{|\Omega|}\sum_{i \in \Omega} {\tilde{p}_i^b} \log(\frac{\tilde{p}_i^b}{\tilde{p}_i^a})
\end{equation}
Note that the above loss only back-propagates to $\theta^a$ and the gradient is stopped for $\theta^b$ (the teacher). Symmetrically, we can switch the teacher and student, and the KD loss for distilling knowledge from $\theta^a$ to $\theta^b$ is denoted as $\mathcal{L}_{KD}(\tilde{P}^b,\tilde{P}^a)$. The overall losses for training $\theta^a$ and $\theta^b$ are:
\begin{equation}\label{eq:overall-2-f1}
		\mathcal{L}_{\theta^a} =
		\lambda\mathcal{L}_{KD}(\tilde{P}^a,\tilde{P}^b) +(1-\lambda)\mathcal{L}_{self}({{P}^a},{\hat{Y}^a})
\end{equation}
\begin{equation}\label{eq:overall-2-f2}
		\mathcal{L}_{\theta^b} = \lambda \mathcal{L}_{KD}(\tilde{P}^b,\tilde{P}^a) + (1-\lambda)\mathcal{L}_{self}({P}^b,\hat{Y}^b)
\end{equation}
where ${L}_{self}$ is a self-training term that uses pseudo labels $\hat{Y}^a$ and $\hat{Y}^b$  to supervise the two networks, respectively. It is implemented by a combination of cross-entropy loss and Dice loss for standard fully supervised learning. $\lambda$ is a hyper-parameter to balance the contribution of self-training and KD.

\section{Experiments and results}\label{sec:experiments}
\subsection{Datasets}\label{sec:experiments:datasets}
We validated our proposed PA-Seg with two applications: 1) Segmentation of vestibular schwannoma from Magnetic Resonance Imaging (MRI) scans, and 2) Segmentation of glioma from MRI scans. 
Note that for both datasets, we abandoned the pixel-level annotations for training images and instead provided point annotations for them. 
For quantitative analysis, we adopted Dice, Average Symmetric Surface Distance (ASSD) between segmentation results and the ground truth to evaluate the segmentation performance. 

\subsubsection{Vestibular Schwannoma (VS) Dataset}
A public VS dataset~\cite{shapey2021segmentation,wang2019automatic} was used for the VS segmentation, and it consists of contrast-enhanced T1-weighted (ceT1) and high-resolution T2-weighted (hrT2) MRI images collected from 242 patients with VS undergoing Gamma Knife stereotactic radiosurgery. In this work we used the hrT2 images for segmentation, with in-plane resolution of approximately $0.5 mm \times 0.5 mm$, in-plane matrix of $384 \times 384$ or $448 \times 448$, slice thickness of $1.0$–$1.5 mm$ and 40-80 slices in a scan. As the segmentation target is relatively small, we cropped and padded if necessary the images to a uniform size of $128 \times 128 \times 48$. We randomly split the dataset into 176, 20 and 46 cases for training, validation and testing, respectively. 
\subsubsection{Brain Tumor Segmentation (BraTS) Dataset}
For glioma segmentation, we used the BraTS 2019 \cite{menze2014multimodal,bakas2017advancing} dataset that was collected from 335 glioma patients across 19 institutions. Each patient was scanned with T1, ceT1, T2 and Fluid Attenuated Inversion Recovery (FLAIR) MRI that were co-registered to the same anatomical template. The images had been skull-stripped and re-sampled to the resolution of $1mm \times 1 mm \times 1 mm$ with size of $240 \times 240 \times 155$. 
In this work, we used the FLAIR images for whole glioma segmentation, and they were randomly split into 244 for training, 27 for validation and 64 for testing, respectively.

\subsection{Implementation Details}
Our framework was implemented in PyTorch and PyMIC\footnote{\url{https://github.com/HiLab-git/PyMIC}}~\cite{Wang2022pymic} on a Ubuntu desktop with an NVIDIA GeForce RTX 3090 GPU, and the code is available online\footnote{\url{https://github.com/HiLab-git/PA-Seg}}. We used the 2.5D U-Net~\cite{wang2019automatic}  as the primary network for the VS dataset due to its anisotropic resolution, and used the 3D U-Net~\cite{ronneberger2015u} as the primary network for the BraTS dataset. 
For each network, we set the channel number in the first block as 16, and it was doubled after each down-sampling layer in the encoder, increasing to 256 at the bottleneck. The patch size was $128 \times 128 \times 48$ and $128 \times 128 \times 128$ for the VS and BraTS datasets, respectively. The batch size was 1 for both datasets. Random cropping and flipping along each of the three axes were used for data augmentation. Stochastic Gradient Descent (SGD) was used for training, with weight decay $3 \times 10^{-5}$ and momentum 0.99. 

For the first stage of PA-Seg, the maximal training epoch $e_{max}$ was set to 300 and 500 for the two datasets, respectively.
The learning rate was initialized as $\eta_0=10^{-2}$, and changed based on a polynomial decay schedule: $\eta = \eta_0 \times (1 - \frac{e}{e_{max}})^{0.9}$, where $e$ is the current epoch. For the CRF loss, $K_{ij}$ was implemented by a single Gaussian kernel with bilateral feature vectors, and the spatial coordinate bandwidth and image intensity bandwidth were 6 and 0.1, respectively. The hyper-parameters $\alpha$ and $\beta$ were set to 1.0 and 0.1, respectively.

In the second stage using SCM,  attention U-Net~\cite{schlemper2019attention} was used as the auxiliary network for both datasets, and it was pre-trained in the same way as the primary network in the first stage.   
For CKD, $T$ was set to 4 according to grid search, and it was chosen by the best final segmentation performance on the validation set. For both datasets, the maximal training epoch was $100$, and the pseudo labels and ``soft" probability maps given by the teacher in $\mathcal{L}_{KD}$ were updated per $20$ epochs for stable training, and the other experimental settings were the same as those in the first stage. We found that after training using SCM, the primary and auxiliary networks had a very close performance, so we only used the primary network for inference on testing images. In addition,  the standard Softmax prediction branch was used for inference at testing time.

\begin{figure}
	\centering
	\includegraphics[width=0.49\textwidth]{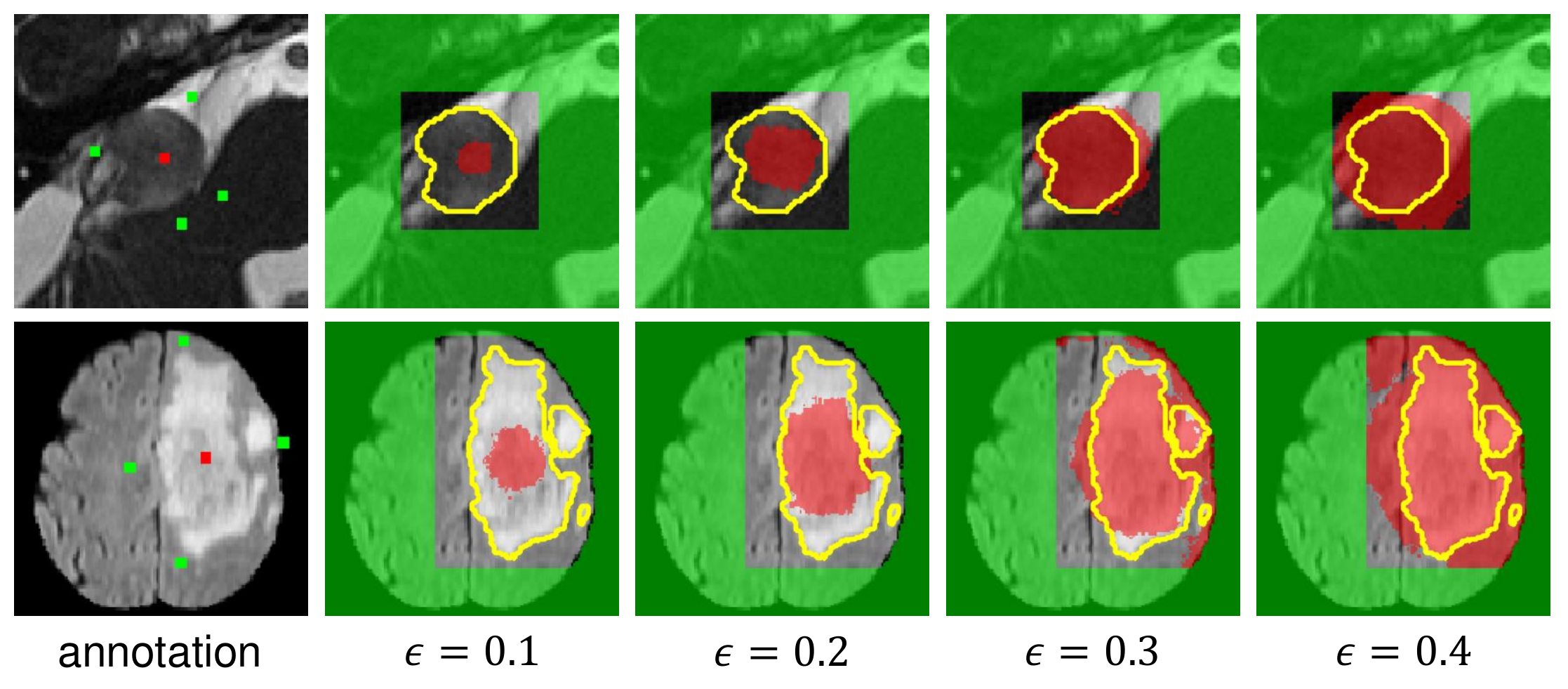}
	\caption{Examples of expanded annotations based on different $\epsilon$ values. The two rows are for VS and glioma, respectively. The yellow curves show the boundaries of full dense annotations. Red and green colors show expanded foreground and background pixels, respectively, and the remaining pixels are unlabeled.}
	\label{fig:threshold-visual}
\end{figure}
\begin{figure}
	\centering
	\includegraphics[width=0.49\textwidth]{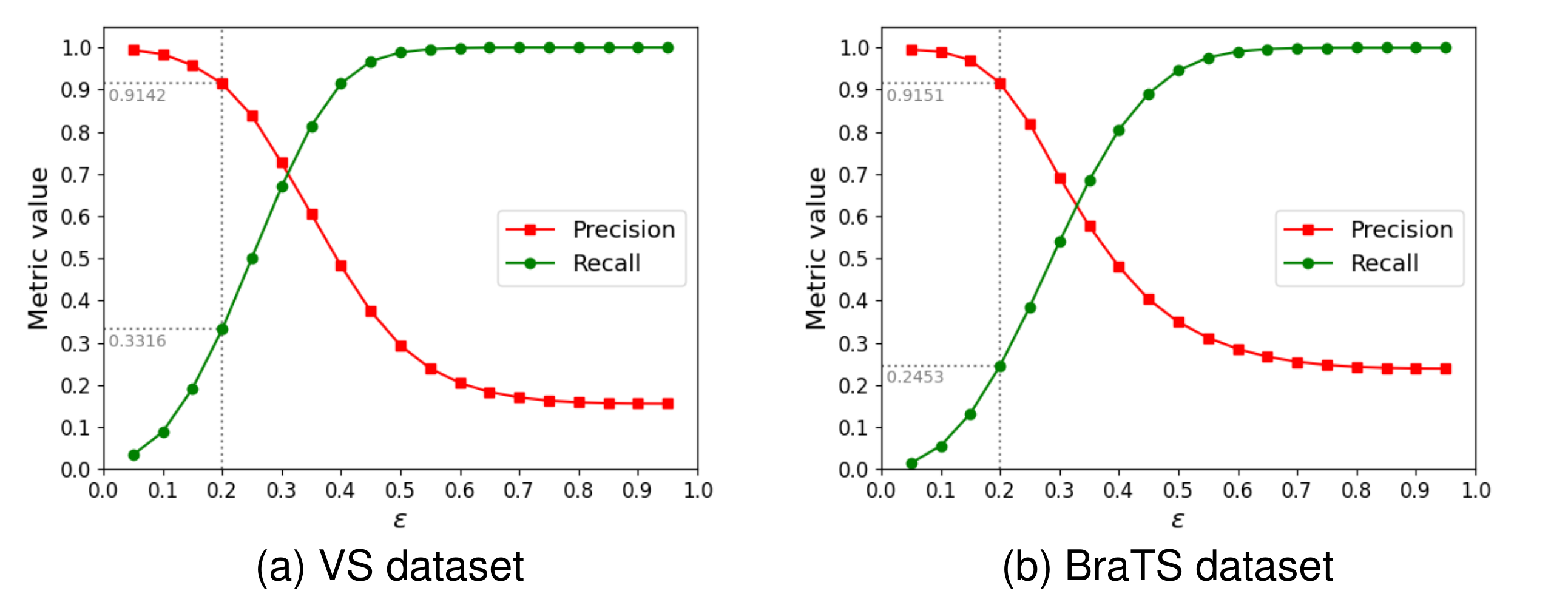}
	\caption{Mean precision and recall between expanded foreground seeds and ground truth masks with different $\epsilon$ values on validation datasets.}
	\label{fig:threshold-curve}
\end{figure}

\subsection{Results of The First Stage}\label{sec:experiments:segmentation}
In this section, we report the results of the initial segmentation model trained in the first stage, i.e., only learning from the expanded seeds with contextual regularization, and the results based on SCM in the second stage will be shown in  \ref{sec:experiments:SCM}.

\subsubsection{Threshold for Seed Expansion}\label{sec:experiments:threshold}
The hyper-parameter $\epsilon$ controls the degree of expansion of the foreground seed. As shown in Fig.~\ref{fig:threshold-visual}, when $\epsilon$ is small (e.g., 0.1), the region of expanded foreground annotation is small, with a high precision and low recall. The foreground seed region grows as $\epsilon$ increases, and it  have a lot of false positives when $\epsilon$ is large (e.g., 0.3 and 0.4), leading to  high recall but low precision. 
To encourage a large number of foreground seeds to provide strong supervision, a high recall is expected. At the same time, to avoid incorrect  annotations in the expanded region, a high precision is desired. Therefore, a proper $\epsilon$ should be selected to balance the   precision and recall. 

We argue that ensuring a high precision is more important than a high recall, as the expanded seeds should not contain many wrong labels that will corrupt the model. The evolution curves of precision and recall of expanded foreground seeds on the validation datasets are shown in Fig.~\ref{fig:threshold-curve}. It can be observed that when $\epsilon=0.2$,  the precision is higher than 0.9 and the recall is 0.33 and 0.25 for VS and glioma, respectively. Using a larger $\epsilon$ leads to a rapid decrease of precision. Therefore, we set $\epsilon = 0.2$ on both datasets. This can also be confirmed from Fig.~\ref{fig:threshold-visual}
where $\epsilon = 0.2$ leads to a large region of expanded foreground seeds with few false positives. 
\begin{figure}
	\centering
	\includegraphics[width=0.49\textwidth]{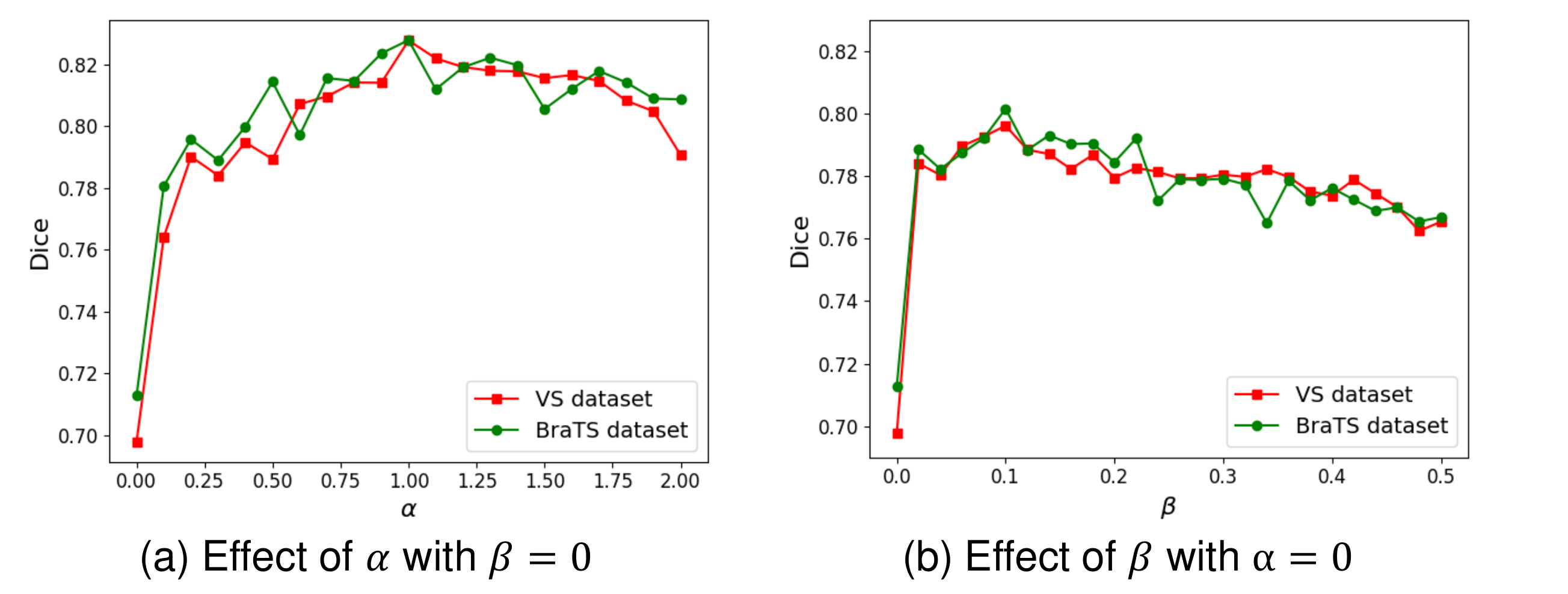}
	\caption{Effect of hyper-parameters $\alpha$ and $\beta$ on  validation datasets.}
	\label{fig:adjust_mCRF_VM}
\end{figure}
\begin{figure}
	\centering
	\includegraphics[width=0.49\textwidth]{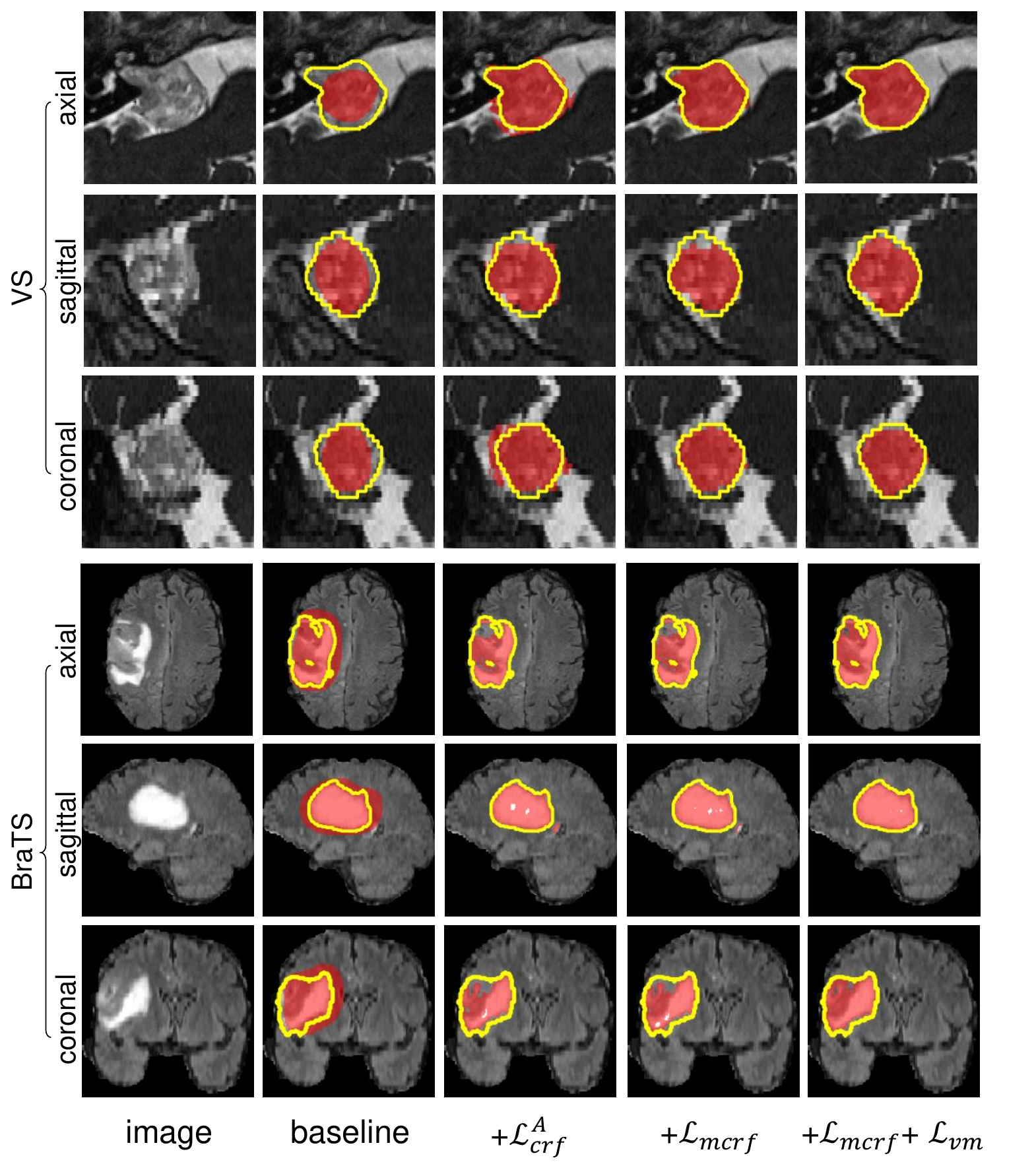}
	\caption{Visual comparison between different losses for learning from expanded partial annotations in the first stage. The yellow curves show ground truth segmentation.}
	\label{fig:mCRF}
\end{figure}

\begin{table}
	\centering
	\setlength{\tabcolsep}{4.0pt}
	\scriptsize
	\caption{Quantitative evaluation of different loss combinations in the first stage of PA-Seg. 
	}
	\label{tab:loss-combination}
	\begin{tabular}{lcccc}
		\toprule
		\multirow{2}{*}{Method} &\multicolumn{2}{c}{VS dataset} &\multicolumn{2}{c}{BraTS dataset} \\ \cline{2-5}
		&Dice & ASSD (mm) &Dice & ASSD (mm)\\ \hline
		full supervision & 0.893$\pm$0.052 & 0.381$\pm$0.219 & 0.877$\pm$0.094 & 2.282$\pm$3.282 \\
		baseline & 0.697$\pm$0.080 & 1.289$\pm$0.537  & 0.718$\pm$0.120 & 4.904$\pm$3.460 \\ \hline
		+$\mathcal{L}^A_{crf}$ & 0.812$\pm$0.090 & 0.740$\pm$0.317& 0.816$\pm$0.152 & 2.894$\pm$2.858 \\
		+$\mathcal{L}_{mcrf}$ & 0.823$\pm$0.095 & 0.673$\pm$0.437 & 0.828$\pm$0.113 & 2.574$\pm$2.698\\
		+$\mathcal{L}_{vm}$ & 0.798$\pm$0.069 &0.756$\pm$0.376 & 0.802$\pm$0.129 & 2.713$\pm$2.595 \\   
		+$\mathcal{L}_{mcrf}$+$\mathcal{L}_{vm}$ & \textbf{0.836$\pm$0.099} & \textbf{0.622$\pm$0.453} & \textbf{0.842$\pm$0.120} & \textbf{2.386$\pm$2.690}\\ 
		\bottomrule
	\end{tabular}
\end{table}
\subsubsection{The Contribution of Each Loss Term}\label{sec:experiments:weakly-contribution}
To validate effectiveness of the multi-view CRF loss and VM loss terms in Eq.~\ref{eq:overall-1}, we first investigated the influence of their weights $\alpha$ and $\beta$ on segmentation performance. We fixed one weight to zero and varied the other, and calculated mean Dice on the validation datasets. The results are shown in Fig.~\ref{fig:adjust_mCRF_VM}. It can be observed that setting $\alpha > 0$ and $\beta > 0$ leads to a large improvement from using the partially supervised loss $\mathcal{L}_{psup}$ alone, respectively. Fig.~\ref{fig:adjust_mCRF_VM} shows that the optimal values of $\alpha$ and $\beta$ are $1.0$ and $0.1$, respectively on both datasets, and the performance does not change much when the weights are around the optimal values. 

Then, we conducted ablation study on the loss terms in Eq.~\ref{eq:overall-1}. The results are shown in Table~\ref{tab:loss-combination}, where the baseline means learning from the expanded annotations only using the partially supervised loss $\mathcal{L}_{psup}$. ``Full supervision" means learning with full annotations based on cross-entropy loss and Dice loss, which serves as the upper bound.
Compared with the baseline, using 2D CRF loss ($\mathcal{L}^A_{crf}$) and VM loss improved the average Dice from 0.697 to 0.812 and 0.798 on the VS dataset, respectively, and they improved the average Dice from 0.718 to 0.816 and 0.802 on the BraTS dataset, respectively. The multi-view CRF loss outperformed the 2D CRF loss on both datasets. The proposed loss combining $\mathcal{L}_{mcrf}$ and $\mathcal{L}_{vm}$ obtained an average Dice of 0.836 and 0.842 on the two datasets, respectively, which outperformed the other variants. A visual comparison of these variants is shown in Fig.~\ref{fig:mCRF}, which demonstrates the effectiveness of our method in learning from partial annotations for 3D segmentation. 

\begin{table}
	\centering
	\setlength{\tabcolsep}{4.0 pt}
	\scriptsize
	\caption{Quantitative comparison between our method in the first stage with state-of-the-art weakly supervised methods. Ours (extreme)
 denotes replacing six background points with extreme points for annotation.}
	\label{tab:sota-weakly}
	\begin{tabular}{lcccc}
		\toprule
		\multirow{2}{*}{Method} &\multicolumn{2}{c}{VS dataset} &\multicolumn{2}{c}{BraTS dataset} \\ \cline{2-5}
		&Dice & ASSD (mm) &Dice & ASSD (mm)\\ \hline
		full supervision & 0.893$\pm$0.052 & 0.381$\pm$0.219 & 0.877$\pm$0.094 & 2.282$\pm$3.282 \\
		baseline & 0.697$\pm$0.080 & 1.289$\pm$0.537  & 0.718$\pm$0.120 & 4.904$\pm$3.460 \\ \hline
		size loss \cite{kervadec2019constrained} & 0.768$\pm$0.087 & 1.007$\pm$0.585  & 0.784$\pm$0.121 & 3.805$\pm$3.555 \\
		DTP \cite{kervadec2020bounding} & 0.780$\pm$0.101 & 0.884$\pm$0.419  & 0.794$\pm$0.138 & 3.696$\pm$4.836 \\
		Deepcut \cite{rajchl2016deepcut} & 0.743$\pm$0.115 & 1.009$\pm$0.525  & 0.755$\pm$0.173 & 7.700$\pm$7.328 \\
		Mumford-Shah \cite{kim2019mumford} & 0.731$\pm$0.084 & 1.128$\pm$0.484  & 0.736$\pm$0.110 & 4.623$\pm$3.438 \\
		EM \cite{grandvalet2004semi} & 0.722$\pm$0.077 &1.183$\pm$0.423 & 0.741$\pm$0.141 & 4.984$\pm$4.742 \\
		TV \cite{javanmardi2016unsupervised} & 0.711$\pm$0.080 &1.302$\pm$0.490 & 0.721$\pm$0.098 & 4.776$\pm$2.986 \\
		ours & \textbf{0.836$\pm$0.099} & \textbf{0.622$\pm$0.453} & \textbf{0.842$\pm$0.120} & \textbf{2.386$\pm$2.690}\\ 
		ours (extreme) & 0.801$\pm$0.068 & 0.723$\pm$0.268 & 0.812$\pm$0.152 & 2.894$\pm$2.858 \\
		\bottomrule
	\end{tabular}
\end{table}

\begin{figure*}
	\centering
	\includegraphics[width=0.95\textwidth]{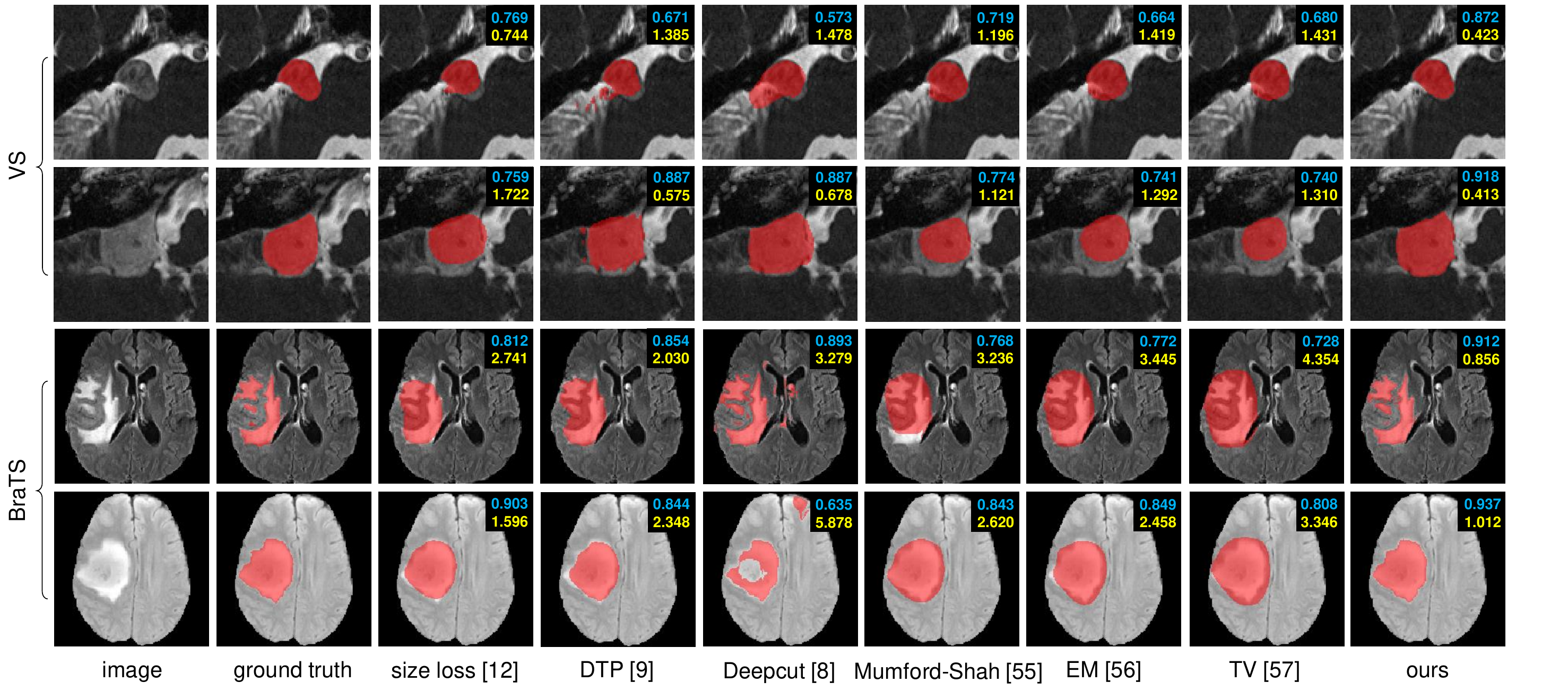}
	\caption{Visual comparison between our method in the first stage and existing weakly supervised segmentation methods. The blue and yellow numbers represent Dice  and ASSD (mm) values, respectively.
	}
	\label{fig:sota-weakly}
\end{figure*}

\subsubsection{Comparison with State-of-the-art Weakly Supervised Learning Methods}\label{sec:experiments:sota-weakly}
The model trained in the first stage of our method was compared with six state-of-the-art weakly supervised methods: 1) size loss \cite{kervadec2019constrained} that employs a global inequality constraint on the volume of the target area, 2) Deep Tightness Prior (DTP) \cite{kervadec2020bounding} that combines topological and emptiness priors with size loss \cite{kervadec2019constrained}, 3) Deepcut \cite{rajchl2016deepcut} that iterates model training and dense CRF post-processing, 4) Mumford-Shah loss \cite{kim2019mumford} that adapts Mumford-Shah energy function in level-set to constrain the segmented region, 5) Entropy Minimization (EM) \cite{grandvalet2004semi} that encourages the model to produce high-confidence predictions for unannotated pixels, and 6) Total Variation (TV) loss \cite{javanmardi2016unsupervised} that penalizes the L1-norm of the gradient of the predicted probability map. The lower and upper bounds used in size loss \cite{kervadec2019constrained} and DTP \cite{kervadec2020bounding} were  0.9 and 1.1 times of the volume of the target, respectively. For DTP~\cite{kervadec2020bounding} and Deepcut~\cite{rajchl2016deepcut} that learn from bounding box annotations, we obtained the bounding boxes based on the same set of six background points as our method with the foreground seed ignored. Note that except for Deepcut \cite{rajchl2016deepcut} and DTP \cite{kervadec2020bounding}, the other compared methods used the same expanded annotations as our method for training. 

\begin{figure}[htb]
	\centering
	\includegraphics[width=0.4\textwidth]{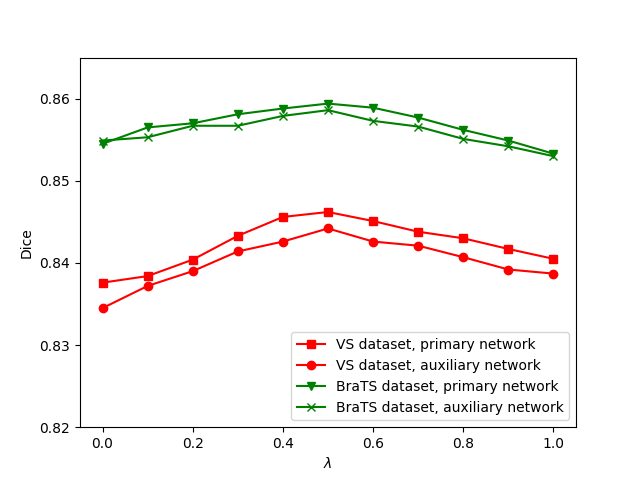}
	\caption{Performance of the primary and auxiliary networks on validation datasets with different $\lambda$ values for SCM in the second stage of PA-Seg.}
	\label{fig:adjust_KDLoss}
\end{figure}

Quantitative evaluation results of these methods are shown in Table~\ref{tab:sota-weakly}. EM~\cite{grandvalet2004semi} and TV~\cite{javanmardi2016unsupervised} performed slightly better  than the baseline. Mumford-Shah loss~\cite{kim2019mumford} and Deepcut~\cite{rajchl2016deepcut} outperformed these two methods, and they achieved an average Dice of 0.731 and 0.743 on the VS dataset, respectively, and 0.736 and 0.755 on the BraTS dataset, respectively.  
DTP~\cite{kervadec2020bounding} achieved the top performance among the existing methods, with Dice of 0.780 and ASSD of 0.884~$mm$ for VS, and Dice of 0.794 and ASSD of 3.696~$mm$ for BraTS. 
Notably, our method largely outperformed the existing weakly supervised methods, achieving Dice and ASSD of 0.836 and 0.622 $mm$ for VS, respectively, and Dice and ASSD of 0.842 and 2.386 $mm$ for the BraTS dataset, respectively.

Fig.~\ref{fig:sota-weakly} shows some qualitative evaluation results. 
Although size loss~\cite{kervadec2019constrained} and DTP~\cite{kervadec2020bounding} outperformed the other existing methods, they have some visible errors nearby the target boundaries, because they only impose some constrains on the volume and topology of the target, without considering the contrast between foreground and background. 
The results of our method have the highest overlap ratio with the ground truth without much false positives and false negatives, indicating the effectiveness and robustness of our method.

\subsubsection{Comparison with Annotation with Extreme Points}\label{sec:experiments:anno_strategy}
As shown in  Fig.~\ref{fig:anno_strategy}, conventional extreme points is less suitable for 3D medical images, as the bounding boxes derived from extreme points in a user-selected slice may not include the entire foreground region in the other slices, which may mislead the model during training. Our proposed point annotations are relaxed to the background in some distance from the exact 3D extreme points, therefore expanding the bounding boxes and reducing the number of false negative pixels. For comparison, we replaced the six background points with extreme points for annotation while reserving one foreground point, which is denoted as ours (extreme). The corresponding results  are shown in the last row of Table~\ref{tab:sota-weakly}. It can be observed that the average Dice dropped from 0.836 to 0.801 on the VS dataset, and 0.842 to 0.812 on the BraTS dataset, respectively.

\subsection{Results of SCM for Noise-robust Learning}\label{sec:experiments:SCM}

\subsubsection{Contribution of Self-training and CKD}\label{sec:experiments:noise-contribution}
\begin{figure*}
	\centering
	\includegraphics[width=0.95\textwidth]{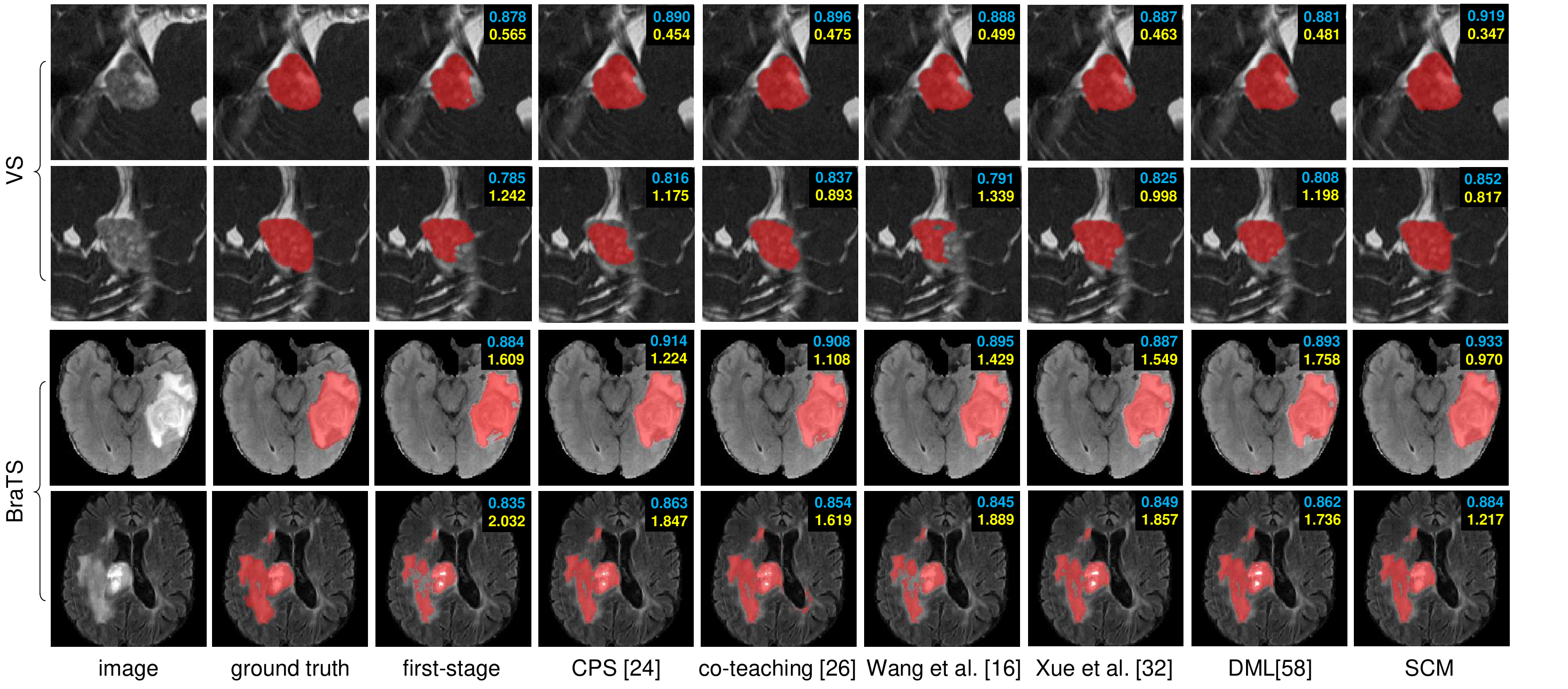}
	\caption{Visual comparison between different noise-robust learning methods on some hard cases  in the second stage of our method. The blue and yellow numbers represent Dice  and ASSD (mm) values, respectively.
	}
	\label{fig:sota-noise}
\end{figure*}
We then used the pre-trained primary and auxiliary models for noise-robust learning based on SCM in the second stage. 
To investigate the influence of the hyper-parameter $\lambda$ in Eq.~\ref{eq:overall-2-f1} and Eq.~\ref{eq:overall-2-f2}, we varied its value from $0.0$ to $1.0$ and calculated the Dice scores on validation datasets, as shown in Fig.~\ref{fig:adjust_KDLoss}.
Note that $\lambda=0.0$  and $\lambda=1.0$ means only using the self-training loss and the CKD loss, respectively. Fig.~\ref{fig:adjust_KDLoss} shows that the optimal value was $0.5$, indicating a combination of the two kinds of losses performed better than using only one of them. 
The last section in Table~\ref{tab:sota-noise} shows 
 a quantitative comparison between $\mathcal{L}_{self}$ ($\lambda=0.0$), $\mathcal{L}_{KD}$ ($\lambda=1.0$) and $\lambda=0.5$ (their combination) on testing datasets. 
Compared with the first-stage model, using $\mathcal{L}_{self}$ and $\mathcal{L}_{KD}$ improved the average Dice from 0.836 to 0.838 and 0.846, respectively on the VS dataset, and from 0.842 to 0.844 and 0.850 on the BraTS dataset, respectively, showing the CKD loss is superior to the self-training loss. SCM ($\lambda=0.5$) further improved the average Dice to 0.852 on the VS dataset and 0.856 on the BraTS dataset, respectively.

\subsubsection{Effectiveness of Asymmetric Architectures for SCM}\label{sec:experiments:effectiveness}
To investigate the effectiveness of using asymmetric architectures for primary and auxiliary networks, we compared it with using the same architecture for the two networks for SCM. The results are shown in Table~\ref{tab:asymmetry}, where Net-A means 2.5D U-Net~\cite{wang2019automatic} and 3D U-Net~\cite{ronneberger2015u} for VS and BraTS datasets, respectively, and Net-B means attention U-Net~\cite{schlemper2019attention}.
It can be observed that when Net-A was used for both of the
primary and auxiliary networks, the Dice improvement from the first stage was 0.006 and 0.004 on the two datasets, respectively. When Net-B was used for both of them, the Dice improvement from the first stage was 0.005 and 0.006 on the two datasets, respectively.
In contrast, when Net-A and Net-B were used for the primary and auxiliary networks respectively, the Dice improvement from the first stage was 0.016 (0.836 to 0.852) and 0.014 (0.842 to 0.856) on the two datasets, respectively, which shows the advantages of using asymmetric network structures for SCM. 

\subsubsection{Performance of Primary Model Compared with Ensemble for Inference}
Fig.~\ref{fig:adjust_KDLoss} shows the performance of the primary and auxiliary models on the validation sets during training. 
We observed that after SCM, the two different networks had a very slight difference, i.e. up to about 0.003 and 0.002 in terms of Dice for VS and BraTS datasets, respectively. 

Besides, we compared inference by the primary model with inference by an ensemble of the primary and auxiliary models, which is denoted as SCM-E ($\lambda=0.5$). The results in Table~\ref{tab:sota-noise} show that the ensemble led to a quite slight improvement from the primary model, i.e., 0.004 and 0.002 in terms of Dice for VS and BraTS, respectively. Therefore, we only used the primary network for inference at test time due to its efficiency and high accuracy.
\begin{table}
	\centering
	\setlength{\tabcolsep}{4.0pt}
	\scriptsize
	\caption{Quantitative evaluation of different noise-robust learning methods in the second stage of PA-Seg. * denotes significant improvement 
 (p-value $<$ 0.05) achieved by our method compared with the first-stage. SCM-E denotes an ensemble of the primary and auxiliary networks.}
	\label{tab:sota-noise}
	\begin{tabular}{lcccc}
		\toprule
		\multirow{2}{*}{Method} &\multicolumn{2}{c}{VS dataset} &\multicolumn{2}{c}{BraTS dataset} \\ \cline{2-5}
		&Dice & ASSD (mm) &Dice & ASSD (mm)\\ \hline
		full supervision & 0.893$\pm$0.052 & 0.381$\pm$0.219 & 0.877$\pm$0.094 & 2.282$\pm$3.282 \\
		first-stage & 0.836$\pm$0.099 & 0.622$\pm$0.453 & 0.842$\pm$0.120 & 2.386$\pm$2.690 \\ \hline
            
		CPS\cite{chen2021semi} & 0.841$\pm$0.089 & 0.595$\pm$0.389 & 0.847$\pm$0.115 & 2.698$\pm$3.138 \\
		co-teaching\cite{han2018co} & 0.840$\pm$0.088 & 0.621$\pm$0.397 & 0.842$\pm$0.124 & 2.844$\pm$3.697 \\
		Wang et al.\cite{wang2020noise} & 0.844$\pm$0.094 & 0.588$\pm$0.437 & 0.845$\pm$0.102 & 2.888$\pm$3.322 \\
		Xue et al.\cite{xue2020cascaded} & 0.843$\pm$0.098 & 0.598$\pm$0.449 & 0.845$\pm$0.107 & 2.483$\pm$3.131 \\
		DML\cite{zhang2018deep} & 0.838$\pm$0.091 & 0.613$\pm$0.402 & 0.843$\pm$0.128 & 2.719$\pm$3.278 \\ \hline
		SCM($\lambda=0$) & 0.838$\pm$0.090 & 0.613$\pm$0.407 & 0.844$\pm$0.114 & 2.872$\pm$3.615 \\
		SCM($\lambda=1$) & 0.846$\pm$0.082 & 0.577$\pm$0.386 & 0.850$\pm$0.114 & 2.241$\pm$2.421 \\
        SCM($\lambda=0.5$) & \textbf{0.852$\pm$0.082}* & \textbf{0.557$\pm$0.399}* & \textbf{0.856$\pm$0.103}* & \textbf{2.204$\pm$2.169} \\
        SCM-E($\lambda=0.5$) & 0.856$\pm$0.079 & 0.543$\pm$0.384 & 0.858$\pm$0.102 & 1.910$\pm$1.511 \\
		\bottomrule
	\end{tabular}
\end{table}

\begin{table}
	\centering
	\setlength{\tabcolsep}{4.0pt}
	\scriptsize
	\caption{Comparison of different network settings for SCM in the second stage of our method. The first section shows results for the first stage. Net-A: 2.5D U-Net~\cite{wang2019automatic} and 3D U-Net~\cite{ronneberger2015u} for VS and BraTS datasets, respectively; Net-B: Attention U-Net~\cite{schlemper2019attention}.}
	\label{tab:asymmetry}
	\begin{tabular}{cccccc}
		\toprule
		\multirow{2}{*}{Primary} & \multirow{2}{*}{Auxiliary} & \multicolumn{2}{c}{VS dataset}  & \multicolumn{2}{c}{BraTS dataset}  \\
		 \cline{3-6}
		& & Dice & ASSD (mm) & Dice & ASSD (mm) \\ \hline
		Net-A & None & 0.836$\pm$0.099 & 0.622$\pm$0.453 & 0.842$\pm$0.120 & 2.386$\pm$2.690 \\
		Net-B & None & 0.831$\pm$0.092 & 0.652$\pm$0.423 & 0.841$\pm$0.107 & 2.381$\pm$2.303 \\ \hline
		Net-A & Net-A & 0.842$\pm$0.088 & 0.598$\pm$0.416 & 0.846$\pm$0.114 & 2.315$\pm$2.507 \\
		Net-B & Net-B & 0.836$\pm$0.091 & 0.628$\pm$0.404 & 0.847$\pm$0.118 & 2.320$\pm$2.153 \\
		Net-A & Net-B & \textbf{0.852$\pm$0.082} & \textbf{0.557$\pm$0.399} & \textbf{0.856$\pm$0.103} & \textbf{2.204$\pm$2.169} \\
		\bottomrule
	\end{tabular}
\end{table}

\subsubsection{Comparison with State-of-the-art Methods for Learning from Noisy Annotations}\label{sec:experiments:noise-sota}

Our SCM was compared with five state-of-the-art methods: 1) CPS~\cite{chen2021semi} using masks predicted by two networks to supervise each other, 2) co-teaching \cite{han2018co} with two networks as well where each network selects samples with low loss values to supervise the other, 3) Wang et al.~\cite{wang2020noise} combining a noise-robust Dice loss with an
adaptive self-ensembling framework, 4) Xue et al.~\cite{xue2020cascaded}  utilizing three networks to select clean samples and correct each other by a joint optimization framework, 
and 5) Deep Mutual Learning (DML)~\cite{zhang2018deep} minimizing Kullback-Leibler divergence between predictions from two networks and cross-entropy loss simultaneously.  For Wang et al.~\cite{wang2020noise} and Xue et al.~\cite{xue2020cascaded}, we used the primary networks with different parameter initialization.
For CPS~\cite{chen2021semi}, co-teaching~\cite{han2018co} and DML~\cite{zhang2018deep}, we used the same two networks as our SCM.
Note that DML~\cite{zhang2018deep} used the same label to supervise the two networks, and we implemented it by taking an ensemble of pre-trained primary and auxiliary  networks to generate the label. In addition, DML does not use temperature scaling to soften the probability maps.

Table~\ref{tab:sota-noise} shows the quantitative evaluation of these methods. Compared with CPS~\cite{chen2021semi}, SCM ($\lambda=1$) replaces the hard labels with the soft ones predicted by another network, and obtained better performance, which demonstrates that the KD-based soft labels are helpful for learning from noisy labels. These existing methods  only achieved marginal improvement over the first-stage model, i.e., 0.008 and 0.005 at most in terms of Dice for VS and BraTS datasets, respectively.
There was almost no improvement for DML~\cite{zhang2018deep}, i.e. 0.002 and 0.001 in terms of Dice on VS and BraTS datasets, because the two networks learned from the same set of labels in DML, which is less effective for avoiding the bias of pseudo labels. 
Our SCM ($\lambda=0.5$) was significantly better than the  compared methods, improving the Dice by 0.016 and 0.014 for the VS and BraTS datasets, respectively. What's more, on the BraTS dataset, our method achieved an average Dice of 0.856 that was just a bit lower than the full supervision method, and it was slightly better than the full supervision method in terms of ASSD (2.204 $mm$ vs 2.282 $mm$). 

Fig.~\ref{fig:sota-noise} shows a comparison of these methods on some hard cases that were inaccurately segmented in the first stage. 
Due to the low contrast and ambiguous boundary of hrT2 images of VS, the first-stage model obtained many false negatives. The existing methods corrected these errors to some extent, but they still resulted in some mis-segmented regions. For the glioma cases with low contrast and abnormal anatomical structures, both the first-stage model and the existing methods had an incomplete segmentation with some holes. Our SCM successfully corrected the mis-segmentation on both datasets, showing its effectiveness on alleviating the impact of noise in the pseudo labels predicted by the first-stage model.

\section{Discussion and conclusion}
Our method decomposes the challenging problem of learning from extremely sparse point annotations into two steps: a first step using contextual regularization of unlabeled pixels
to train an initial model, and a second step using SCM to learn from noisy pseudo labels.  The contextual regularization may also be applicable for other scenarios where incomplete annotation exists, such as semi-supervised segmentation~\cite{yu2019uncertainty,luo2021semi} and learning from scribbles~\cite{luo2022scribble}. The SCM is a general framework for noisy-label learning, and it may also be applied to learning from non-expert annotations~\cite{wang2020noise,han2018co}.

Our SCM shares some spirit with Deep Mutual Learning (DML)~\cite{zhang2018deep}, but they have substantial differences. First, DML aims to avoid a pre-trained teacher to improve the performance of a CNN for image classification tasks learning from clean labels. In contrast, our SCM is proposed to deal with noisy labels for 3D medical image segmentation. Second, the same set of fixed labels are used to supervise the two CNNs in DML, but the two CNNs in SCM learn from different sets of labels (each network's prediction for self-training) that are updated after some epochs. Thirdly, DML does not use temperature scaling to soften the probability maps during knowledge distillation, while our method uses soft predictions during distillation to alleviate the effect of noisy labels. Table~\ref{tab:sota-noise} shows that our method largely outperformed DML in learning from noisy labels for segmentation. 

We noticed that Patel and Dolz~\cite{patel2022weakly} also applied KD to weakly supervised medical image segmentation. That work is different from ours in several aspects: First, Patel and Dolz~\cite{patel2022weakly} learned from slice-level annotations for segmentation, while our method learns from point annotations. Second, Patel and Dolz~\cite{patel2022weakly} used networks based on the same architecture to deal with different input modalities with consistency regularization on the class activation maps, while our method uses asymmetric architectures for the two networks to learn from noisy pseudo labels. Thirdly, Patel and Dolz~\cite{patel2022weakly} applied KD to the classification outputs without temperature scaling, while our method applies KD to segmentation predictions that are softened by temperature scaling.


In conclusion, we propose a two-stage weakly supervised 3D segmentation framework PA-Seg using very sparse point annotations, i.e., six and one points for the background and foreground, respectively.  In the first stage, the single foreground  seed is expanded based on geodesic distance transform, and a model is trained with contextual regularization of unannotated pixels based on multi-view CRF and variance minimization.  In the second stage,  a new self and cross monitoring method based on cross knowledge distillation is proposed to learn from noisy pseudo labels generated by the initial model. Experiments on two datasets demonstrated that our model in the first stage already outperformed several existing methods for weakly supervised segmentation. In the second stage, our SCM can further improve the model's performance by robustly learning from noisy labels, which also outperformed existing noise-robust learning methods. Our framework substantially reduces the annotation cost for training high-performance 3D medical image segmentation models. 
\bibliographystyle{IEEEtran}
\bibliography{myref}

\end{document}